\documentclass[10pt,twocolumn,letterpaper]{article}

\usepackage{cvpr}              
\definecolor{cvprblue}{rgb}{0.21,0.49,0.74}
\usepackage[pagebackref,breaklinks,colorlinks,allcolors=cvprblue]{hyperref}

\usepackage{booktabs, tabularx}
\usepackage{multirow}
\usepackage[table]{xcolor}
\usepackage{makecell}
\usepackage[font=scriptsize, labelfont=scriptsize]{caption}

\def\paperID{18572} 
\def\confName{CVPR}
\def\confYear{2026}

\title{Face, Whole-Person, and Object Classification in a Unified Space Via The Interleaved Multi-Domain Identity Curriculum}

\author{
Thomas M Metz \quad Matthew Q Hill \quad Alice J O'Toole \\
School of Behavioral and Brain Sciences, The University of Texas at Dallas \\
Richardson, Texas, USA \\
{\tt\small thomas.metz@utdallas.edu, matthew.hill@utdallas.edu, otoole@utdallas.edu}
}

\newlength\floatmargin
\newlength\capmargin
\begin{document}

\maketitle

\section{Abstract}

Vision foundation models can perform generalized object classification in zero-shot mode, and  face/person recognition when they are fine-tuned. However, fine-tuned models suffer from catastrophic forgetting. We create models that perform four tasks (object recognition, face recognition from high- and low-quality images, and person recognition from whole-body images) in a single embedding space---without incurring substantial catastrophic forgetting. To accomplish this, we introduce two variants of the Interleaved Multi-Domain Identity Curriculum (IMIC): a gradient-coupled, interleaving training schedule that fine-tunes a foundation backbone simultaneously on all four tasks. The IMIC method proved effective with three foundation model bases: DINOv3, CLIP, and EVA-02. Two of these (EVA-02 and CLIP) performed comparably with domain experts on all four tasks concurrently and were more accurate than humans at multi-tasking across face, body, and object datasets. Further, we demonstrate that our approach does not substantially harm out-of-distribution generalization, thus maintaining a key property of foundation models. 
Analysis of the most accurate model variants (EVA-02 + IMIC A and B) showed linearly separable representations of the four tasks in the unified embedding space, but with substantial sharing of features across tasks. Fewer than 100 PCs calculated from any one task could perform all other tasks with nearly zero performance degradation.

\section{Introduction}

The human visual system performs object and person identification tasks in a seemingly integrated way.   Images of the visual world are projected onto the retina through eye-movements that center these projections on points of interest in the environment. In real-world viewing conditions, faces, bodies, and objects are examined in quick succession. These transient retinal images are  processed first in early visual areas that are organized topographically by spatial position on the retina. As information about the image proceeds to downstream neural areas, brain representations become less topographically organized and more specialized by category. At a higher level of visual analysis, the inferior temporal (IT) cortex plays a critical role in visual recognition \cite{DiCarlo2012VisualRecognition}. In IT cortex, different brain regions  specialize in processing objects (LOC \cite{grill2001lateral}), faces (OFA \cite{pitcher2007tms}, FFA \cite{kanwisher2006fusiform}), and bodies (EBA \cite{astafiev2004extrastriate},  FBA \cite{peelen2005selectivity}). These regions appear to be domain-specific, with cortical processing of different categories (face, body, and object) located at different, but semi-adjacent, places in IT cortex.

An open question in visual neuroscience is how the brain can learn highly specialized  representations of faces, bodies, and objects from a diverse and variable input stream. These recognition tasks  differ in fundamental ways. Object recognition is a {\it basic-level categorization task} \cite{mervis1981categorization} in which items  are classified primarily according to 3D shape (e.g., table vs.\ chair). This task requires  robustness to exemplar-based shape variation (e.g., desk vs.\ table lamp). Face recognition is a {\it within-category discrimination task} in which individual items in the category of ``faces'' must be  separated by identity. This requires fine-grained distinctions in facial shape and reflectance  within a head shape that is roughly constant across individuals.   Body recognition falls  between face and object recognition.  Bodies in the real world (e.g., variably clothed) are not sufficiently unique to discriminate by identity, but nor are they  sufficiently categorical to separate  into discrete  shape categories. 

Low-resolution face recognition constitutes yet an additional ``hidden'' task embedded in person recognition from whole body images. Faces are commonly (though not always) present in  whole-person images. However, the identity information in these more distantly viewed faces is degraded. That said, the performance of long-term body recognition algorithms declines substantially when faces are obscured in whole person images \cite{metz2025dissectinghumanbodyrepresentations}. It is likely that aspects of the head shape, facial tone, facial hair, and hair style all provide useful cues for person identification from whole body images, both by humans and machines. 
   
The divergent requirements of face (high- and low-resolution), body, and object recognition pose challenges for algorithms aimed at mastering all four tasks simultaneously. Accordingly, computer vision researchers have traditionally focused separately  on  domain-specific recognition tasks. These specialized tasks include face recognition (commonly trained using large-scale datasets, such as Casia-WebFace \cite{yi2014learningfacerepresentationscratch}, VGGFace2 \cite{cao2018vggface2datasetrecognisingfaces}, and WebFace \cite{zhu2021webface260mbenchmarkunveilingpower}); whole-person recognition (trained with moderately large-scale clothes-change datasets, such as DeepChange \cite{xu2023deepchange} and the BRIAR Research Set, BRS, \cite{cornett2023expanding,jager2025expandingbriardatasetcomprehensive}), and object classification (commonly trained using datasets like ImageNet-1k \cite{ILSVRC15} and ImageNet-21k \cite{ridnik2021imagenet21kpretrainingmasses}). 

By contrast, vision foundation models, such as Contrastive Language-Image Pretraining (CLIP) \cite{radford2021learningtransferablevisualmodels} and Distillation with No labels (DINOv2) \cite{oquab2024dinov2learningrobustvisual}  have been designed to  generalize  across a range of vision tasks. These models are very large and are trained on massive and diverse image data sets. Operating in ``zero-shot'' learning mode (i.e., with no additional training), foundation models have been surprisingly good at tasks for which they are not explicitly trained (see \cite{radford2021learningtransferablevisualmodels}, \cite{Fang_2024}, \cite{siméoni2025dinov3}). However, for face recognition, which requires fine-grained, identity-based distinctions, zero-shot learning applied to vision foundation models performs poorly \cite{chettaoui2025froundationfoundationmodelsready,luu2024clip}.

Moving beyond zero-shot mode, extensive training of foundation models provides an especially good initial state for fine-tuning with a domain-specific transfer task (e.g., face recognition, body recognition \cite{chettaoui2025froundationfoundationmodelsready, metz2025earlybirdidentifiesworm}). Specifically, foundation models fine-tuned with few training epochs and far less domain data can perform at substantially higher levels of accuracy than non-foundation models extensively fine-tuned with large quantities of domain-based data \cite{metz2025earlybirdidentifiesworm}. 

In this work, we start with a pre-trained foundation model and apply multitask learning to produce a network with a single representation space capable of performing four distinct visual classification tasks: object classification, face recognition from high-quality images, face recognition from low-quality images, and whole-person recognition. Our goal was to produce a model that performs as well as (or nearly as well as) domain expert models on these four tasks concurrently, while avoiding catastrophic forgetting of objects. We also aimed for a model that could compete with the multi-tasking skills of humans for face, body, and object recognition. To that end, we introduce a novel multitask learning curriculum that interleaves exemplars of the different types of images during training. Two foundation model bases, trained with the interleaved curriculum achieved these  goals. To our knowledge, these are the first completely unified models capable of performing face, whole-body, and object recognition at, or above, domain specialized models and humans.

\section{Contributions}

\begin{enumerate}
    \item We provide the first work that demonstrates that multi-task learning can be effectively utilized to solve ``intra'' class and ``inter'' class classification problems in a single embedding space.
    \item We develop a novel human-inspired  Interleaved Multi-Domain Identity Curriculum (IMIC), a curriculum to unify the training of face, body, and object classification.
    \item We demonstrate that the IMIC can be used to fine-tune existing foundation models (EVA-02 Large, CLIP, and DINOv3) to perform fine-grained face and person recognition, while continuing to perform well on object classification. We show that performance remains strong on out-of-distribution tasks.
    \item We demonstrate that  our interleaved training framework, when applied to EVA-02 Large and CLIP, produces the first unified models capable of performing face, body, and object classification at a super-human level.
    \item We analyze the joint multi-task embedding space and demonstrate that tasks occupy linearly separable subspaces. Despite this, we demonstrate that the tasks heavily share features, such that any task can be performed at near full accuracy, using fewer than the 100 most important principal components extracted from the subspace of any other task.

\end{enumerate}

\section{Related Work}

\subsection{Vision Foundation Models}

Vision Foundation Models are extensively pretrained on a broad range of image data, often in a self-supervised or semi-supervised fashion.  Contrastive Language-Image Pretraining (CLIP) was the first vision foundation model. It was trained to maximize the cosine similarity between image and text pairs using a dataset of 400 million such pairs \cite{radford2021learningtransferablevisualmodels}. CLIP achieved a 0-shot accuracy of $76.2\%$ on ImageNet-1k, a level nearly equivalent to that of networks specialized for ImageNet classification. 
At around the same time, DINOv2 offered an unsupervised training method  based on a student and teacher model approach \cite{oquab2024dinov2learningrobustvisual}. Specifically, DINOv2 trains a student network to produce representations similar to the teacher network by using strong augmentations based on the teacher network's global view. This achieves over $82\%$ Rank 1 accuracy on ImageNet-1k with  linear probing. A more recent foundation model, EVA-02, applies innovations from natural language processing combined with masked-image-modeling \cite{Fang_2024}. Using an EVA-CLIP teacher, the model achieves strong performance across a range of vision tasks including object classification, instance segmentation, and semantic segmentation. Notably, EVA-02  surpasses $90\%$ rank 1 accuracy  on ImageNet-1k with only 304 million parameters.

Despite the success of vision foundation models, a combination of coarse semantic features and catastrophic forgetting still limit the ability of these models to perform both object classification and more discriminative within-category person identification at human levels. Out of the box, pre-trained CLIP and DINOv2 models perform face identification well above chance, but fall short of dedicated face identification networks---especially in challenging unconstrained scenarios \cite{chettaoui2025froundationfoundationmodelsready}.  Foundation models that are fine-tuned achieve competitive performance, even with limited domain data \cite{chettaoui2025froundationfoundationmodelsready}. However, in these fine-tuned cases, foundation models still suffer from catastrophic forgetting. Work in continual learning suggests that direct fine-tuning can substantially degrade the zero shot accuracy of the CLIP \cite{ding2022dontstoplearningcontinual} and DINOv2 \cite{bafghi2024parameterefficientfinetuningselfsupervised} models.

To some degree, more advanced continual learning techniques have managed to offset catastrophic forgetting when fine-tuning vision foundation models. For example, Block expansion \cite{bafghi2024parameterefficientfinetuningselfsupervised} performs adaptation with only  minimal performance degradation in the pre-training domain. Additionally, Learning Without Forgetting via Replayed Vocabulary (VR-LwF) applies a replaying-based approach to CLIP, offsetting catastrophic forgetting while fine-tuning on new classes \cite{ding2022dontstoplearningcontinual}. Although effective, these methods have not been applied to both recognition of humans and objects. Consequently, a network capable of performing both tasks has not yet been proposed.

\subsection{Multi Task Learning With Interleaving}

Interleaved learning refers to training with a curriculum in which the skills or topics are presented in alternating sequences. This is opposed to learning with a sequential curriculum in which an entire skill/topic is presented before moving to the next skill/topic. When presented with a series of skills to learn, human long-term memory benefits from interleaved (spaced) presentation of items from different categories \cite{kornell2008spacing}---even though it is initially more difficult to learn in this way \cite{bjork2011desirable}. Thus, it can be advantageous to introduce ``desirable'' difficulties into the learning process \cite{bjork2011desirable}. People learn information more permanently when learning is made more difficult by interleaving the presentation of categories to learn. For example, when studying math, interleaving problems reduces student performance on day of learning, yet doubles test scores the next day \cite{taylor2010interleaved}. 

Most machine learning methodologies rely on a blocked training approach, utilizing pre-trained weights to transfer from task A to task B; however, interleaved training has shown promise in a small number of MultiTask learning works. In a multi-task learning simulation study, interleaved training consolidates knowledge from different tasks, helping to offset catastrophic forgetting \cite{52337}. Moreover, in a human-inspired learning approach, interleaved multi-task learning with energy-modulated learning progress (IMTL-EMLP) consistently outperforms blocked learning in effect-prediction tasks executed by a simulated manipulator robot \cite{Say_2025}. 

Interleaved training has not yet been applied to training a single network to perform both object- and person-identification tasks concurrently. Here, we apply it to pre-trained foundation models to produce networks capable of performing four distinct visual classification tasks: face recognition from high-quality images, face recognition from low-quality images, whole-person recognition, and object classification.

\section{Methods}

\subsection{Pre-Training}

Our models are built on three pre-trained foundation models: EVA-02 (L) \cite{Fang_2024}, CLIP (L) \cite{radford2021learningtransferablevisualmodels}, and DINOv3 (L) \cite{siméoni2025dinov3}. In the rest of this paper, when we refer to EVA-02, CLIP, and DINOv3 we are referring to the large variants of these models. Training for all models includes large scale pre-training and self-supervised or semi-supervised learning signals. 


\subsection{Interleaved Multi-Domain Identity Curriculum - Balanced (IMIC-B):}
\label{Cirriculum}

We propose a novel curriculum called Interleaved Multi-Domain Identity Curriculum - Balanced (IMIC-B) to unify four recognition tasks (object, HQ face, LQ face, whole-person) in one embedding space. In IMIC-B, we implement a gradient mixing strategy whereby, for a gradient accumulation factor $g = taskCount * t$, $t$ batches are sampled for each sub-recognition task (taskCount = 4 in this work: Object, Body, HQ Face, LQ Face). Gradients across $g$ batches are summed and backpropagated as a single supervisory signal. IMIC-B enforces equality of task batches per backpropagation, but allows for unequal task dataset sizes by refreshing the data loader for a sub-task when all instances of that dataloader have been utilized. Under IMIC-B, an epoch completes when all data sources are exhausted. 

Rather than deriving special handling of between- and within-class classification, IMIC-B imposes no prior structure on the optimization beyond what is inherited from the fine-tuned weights. By treating all labels equivalently, the underlying structure of the embedding space derives solely from the optimization.

IMIC-B has an undesirable property for practical implementation: each task should have a roughly similar dataset size. If this property is not met, the model may train on a diverse set from one task and a limited set from another, potentially producing skewed performance. We propose IMIC-A as a solution.

\subsection{Interleaved Multi-Domain Identity Curriculum - Adaptive (IMIC-A):}

\begin{figure}[h]
    \centering
    \includegraphics[scale=0.5]{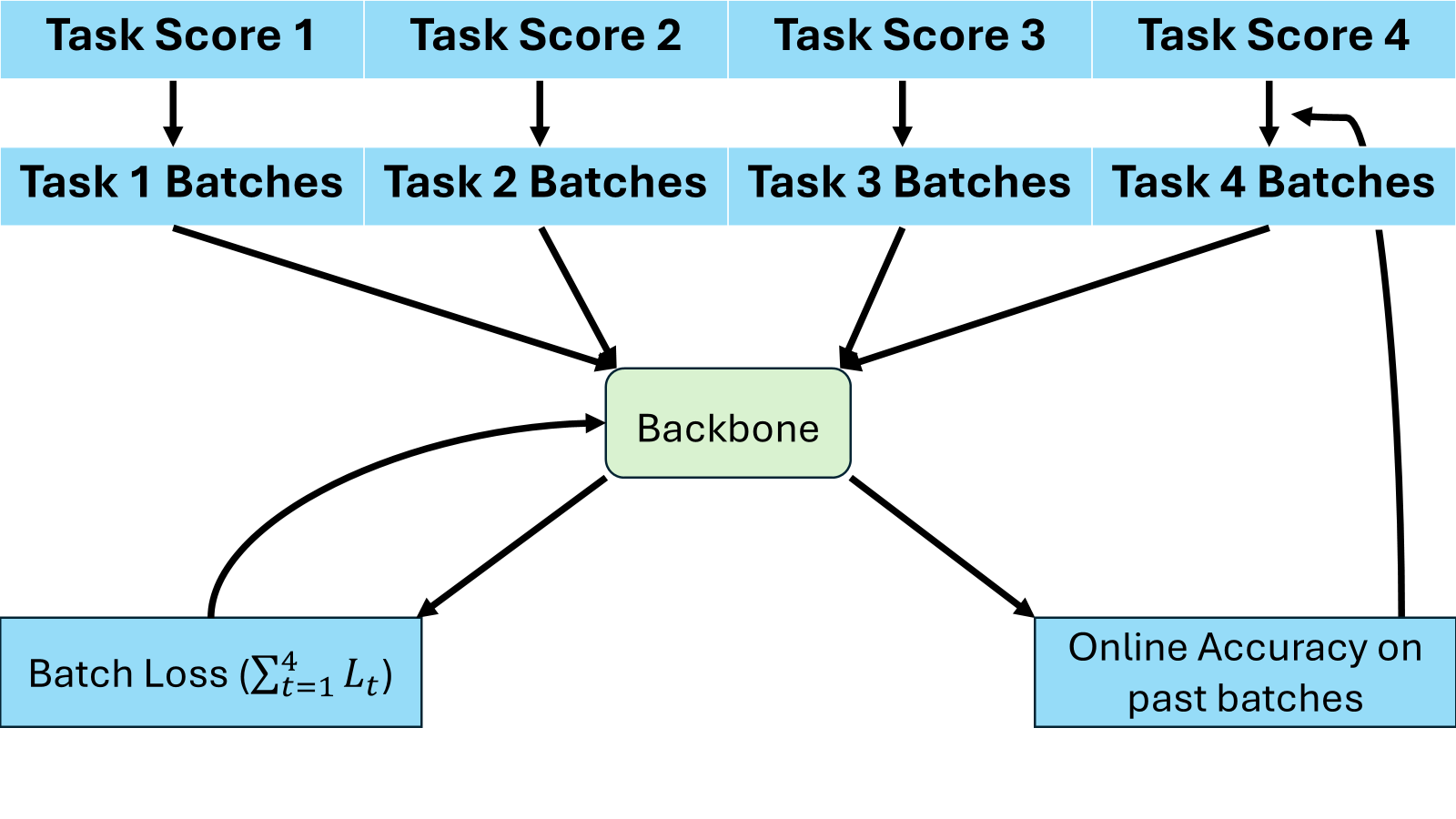}
    \caption{In the IMIC-A training procedure, the batch loss is aggregated as the sum of task losses. All tasks share a common loss function (Triplet Loss in our experiments). Online accuracy on prior training samples for each task informs the frequency at which that task is sampled in future training steps.}
    \label{fig:training-procedure}
\end{figure}

IMIC-A (Figure \ref{fig:training-procedure}) extends the curriculum strategy introduced in IMIC-B by incorporating prior model performance to adaptively determine the number of task batches used per backpropagation step. This approach allows the training process to maintain an interleaved structure while easing requirements on the amount of data that must be available for tasks the model is already good at.

\paragraph{Task Batch Allocation.}

Let the total number of task batches used per backpropagation step be denoted as
\[
T = \sum_{i=1}^{N} t_i ,
\]
where $t_i$ is the number of batches allocated to task $i$, and $N$ is the number of tasks considered (e.g., $N = 4$ in our experiments).

\paragraph{Performance Improvement.}

For each task $i$, let $G_i$ denote the target (or goal) accuracy, and let $\text{Metric}_{\text{prev}}$ and $\text{Metric}_{\text{curr}}$ represent the model’s performance metric on that task before and after the most recent update, respectively. We define the \emph{relative improvement} in performance for task $i$ as: 
\[
R_i = \frac{\lvert \text{Metric}_{\text{curr}} - \text{Metric}_{\text{prev}} \rvert}{\text{Metric}_{\text{prev}}} \cdot \frac{1}{t_i} + \epsilon ,
\]
where $\epsilon$ is a small constant added for numerical stability. This quantity measures the efficiency of performance gains relative to the number of task batches used.

\paragraph{Distance from Goal.}

We define the \emph{distance to the target accuracy} for task $i$ as
\[
D_i = \frac{G_i - A_i}{G_i},
\]
where $A_i$ denotes the current accuracy of the model on task $i$ and $G_i$ denotes the goal accuracy. A larger $D_i$ indicates that the model remains further from the desired performance level. To address the potential issue of no batches being sampled per a task and step, we maintain a log of prior accuracies and refer to the most recent previous accuracy in this case.

\paragraph{Task Scoring and Sampling.}

A task’s \emph{difficulty score} is then computed as
\[
S_i = \frac{D_i}{R_i}.
\]
Tasks that are both far from their goal accuracy (large $D_i$) and showing little recent improvement (small $R_i$) will thus receive higher scores. The scores are subsequently normalized into a probability distribution over tasks:
\[
p_i = \frac{S_i}{\sum_{j=1}^{N} S_j}.
\]

To ensure that no task becomes underrepresented, IMIC-A applies a two-stage normalization process that enforces a minimum sampling threshold, effectively guaranteeing a non-zero lower bound on $p_i$ (set to .05 in our experiments). Finally, task batches for each backpropagation step are sampled in proportion to these adjusted probabilities, thereby adapting the curriculum dynamically based on ongoing model performance.


\begin{table*}[tbhp]
\scriptsize
\centering
\setlength{\tabcolsep}{5pt}
\renewcommand{\arraystretch}{1.2}
\begin{tabular}{|l|c|c|cc|cc|cc|cc|ccc|}
\hline
\multirow{2}{*}{\textbf{Model}} &
\multicolumn{1}{c|}{\textbf{LFW}} &
\multicolumn{1}{c|}{\textbf{CP-LFW}} &
\multicolumn{2}{c|}{\textbf{Market-1501}} &
\multicolumn{2}{c|}{\textbf{ImageNet-1k}} &
\multicolumn{2}{c|}{\textbf{ImageNet-A}} &
\multicolumn{2}{c|}{\textbf{BTS}} &
\multicolumn{3}{c|}{\textbf{TinyFace}} \\
\cline{2-14}
&
\textbf{Acc (\%)} &
\textbf{Acc (\%)} &
\textbf{R@1} & \textbf{T@F$10^{-3}$} &
\textbf{Top-1} & \textbf{Top-5} &
\textbf{R@1} & \textbf{R@5} &
\textbf{R@1} & \textbf{T@F$10^{-3}$} &
\textbf{R@1} & \textbf{R@5} & \textbf{R@10} \\
\hline

\hline
EVA02 + IMIC-A  & 98.47  & \textbf{87.63} & 98.2 & \textbf{79.66} & \textbf{89.09} & \textbf{96.68} & \textbf{85.03} & \textbf{91.53} & 95.1 & 81.8 & \textbf{71.00} & \textbf{79.24} & \textbf{81.70} \\
EVA02 + IMIC-B & 98.80 & 85.93 & 98.07 & 74.16 & 89.11 & 96.55 & 85.37 & 91.36 &  \textbf{96.93} & \textbf{84.21} &  69.02 & 76.88  &  79.37 \\
CLIP + IMIC-A & 98.23 & 80.02 & 98.28 & 64.22 & 84.32 & 95.72 & 64.41 & 81.37 & 95.48 & 80.57 & 68.05 & 76.02 & 78.62\\
CLIP + IMIC-B & \textbf{99.12} & 84.83 &98.28 & 68.89 & 82.85 & 95.18 & 62.25 & 80.16 & 95.13 & 82.75 & 68.32 & 75.99 & 78.51\\
DINOv3 + IMIC-A & 92.25 & 76.47 & 98.27 & 58.58 & 85.04 & 95.52 & 58.99 & 74.61 & 85.96 & 57.99 & 68.24 & 76.21 & 79.29\\
DINOv3 + IMIC-B & 97.02 & 76.68 & \textbf{98.43} & 55.99 & 85.71 & 95.54 & 77.39 & 86.15 & 90.08 & 71.43 & 63.49 & 73.31 & 77.15\\

\hline
EVA02 \cite{Fang_2024} & 88.80  & 67.03 & 87.77 & 10.51 & \textbf{89.89} & \textbf{96.41}& \textbf{85.81} & \textbf{91.99} &26.67 & 04.58 & 46.78 & 57.91 & 61.78 \\
CLIP \cite{chen2020simple} & 96.85 & 79.28 & 87.23 & 09.94 & 79.01 & 93.70 & 59.73 & 80.35 & 85.00 & 12.07 & 38.22 & 48.23 & 53.65 \\
DINOv3 & 85.40 & 62.65  & 95.69 & 13.09 & 77.99 & 91.93 &32.31 & 55.32 &  28.61 & 04.64 & 48.42 & 57.67 & 61.94 \\
AIMV2 \cite{fini2025multimodal} & 87.60 & 67.83  & 94.18 & 16.19 & 80.10 & 93.41 & 42.36 & 65.67 & 54.99 & 05.57 & 30.55 & 41.47 & 46.03 \\

\hline
PetalFace\cite{narayan2024petalface} & \textbf{99.6} & \textbf{93.18} &-- & -- & -- & --& -- & -- & -- & -- & \textbf{75.72} & 78.86 & \textbf{81.70} \\
Echo-Bid\cite{metz2025earlybirdidentifiesworm} & -- & --& 98.01 & 69.85 & --& -- & -- & -- & 95.90 & 84.08 & -- & -- & -- \\
Human & 98.3\cite{6996296}--99.20\cite{5728825} & 81.21 &
93.5& -- & -- &
94.9\cite{ILSVRC15} & -- & -- & -- & -- & -- & -- & -- \\

\hline
\end{tabular}
\vskip .15 cm
\caption{Model Performance: IMIC-trained foundation models (top section); zero shot foundation models (middle), domain-specific models and humans (bottom). Top performance among the models in the first 2 sections is bold-faced. Domain models are bold-faced when they exceed this max.
IMIC-trained
are competitive with or surpass domain experts for body, object, and LQ-Face.}
\label{tab:unified_human_baselines}
\end{table*}

\subsection{Training Datasets}

In IMIC-B, all datasets were randomly sampled to roughly balance their size with the smallest training set (low quality face images). IMIC-A used the complete datasets and relied on model performance to dictate the number of samples drawn from each set. 

{\it Low Quality Face  (LQ Face) Identification}.
Low-quality face data consisted of a combination of TinyFace \cite{cheng2019lowresolutionfacerecognition} and faces detected using an MTCNN detector \cite{8110322} on the following person-identification data: DeepChange \cite{xu2023deepchange}, BRS \cite{cornett2023expanding}, KWBRC \cite{KWBRC}, UAVHuman\cite{li2021uavhuman} , MARS\cite{zheng2016mars}, MSMT17 \cite{wei2018persontgan}, Market-1501 \cite{zheng2015scalable}, PRCC \cite{yang2019prcc}, MEVID, and STR-BRC 1. IMIC-B sampled $100\%$ of this data.

{\it High Quality Face  (HQ Face) Identification}.
WebFace-4m was used for High Quality face identification data. IMIC-B sampled $23.3\%$ of the available training data.

{\it Whole Person Identification}.
The training data for this task was a combination of three unconstrained face datasets that were previously successful in adapting foundation models to whole person identification \cite{metz2025earlybirdidentifiesworm}. These are: DeepChange \cite{xu2023deepchange}, Kitware BRC \cite{KWBRC}, and BRS 1--6 \cite{cornett2023expanding}. IMIC-B sampled $100\%$ of the available data.

{\it Object Classification}.
ImageNet-1k \cite{ILSVRC15} was used for the object classification training data. IMIC-B sampled $80\%$ of the available training data.

\subsection{Test Datasets}

We tested performance with two HQ Face datasets 
(LFW \cite{huang2008labeled}, CP-LFW \cite{zheng2018cross}) and one LQ Face dataset (TinyFace, \cite{cheng2019lowresolutionfacerecognition}).
For bodies/people, we tested on 
(Market-1501), a re-identification benchmark with no clothes change and on a
102-identity subset of the highly challenging 
Briar Test Set, \cite{cornett2023expanding}, which includes
clothes change and images from a wide range of distances and with pitch variation.

\subsection{Implementation Details}

All models were trained using small batch sizes of 40 with a low learning rate ($3e^{-5}$). Models were trained with triplet loss with batch hardest positive and negative mining. 4 positive samples per ID were utilized and a large margin of .35 was set. Both models were trained using an Adam optimizer \cite{kingma2017adammethodstochasticoptimization} with weight decay of $1e^{-6}$.
\vspace{-0.5 em}
\section{Results}

Since there are no previously available  models that provide a unified solution to the diverse tasks we consider (HQ face, LQ face, whole-person identification, and object classification), we compare our models against three other categories of baseline:  zero-shot foundation models, domain-specific expert models, and human performance (where comparable measures were available).   
In addition to standard measures of performance on each of the tasks
(e.g., accuracy, rank measures, and True Accept Rate at False Accept Rate $10^{-3}$),
we introduce two variants  of a multi-tasking measure that 
capture cross-task performance. The first variant is computed as average relative difference from the maximum performance for each metric and dataset. This allows for holistic comparison with domain experts. The second variant is computed as average relative difference from human performance for each metric and dataset for which a human benchmark is available. 
This allows holistic comparison with human multi-task performance. Table \ref{tab:unified_human_baselines} shows the datasets and measures that have a human benchmark.
\begin{equation}
\text{MultiTask}^{expert}_{i} = \frac{1}{m} \sum_{j=1}^{m} \frac{a_{ij} - max(a)_{j} }{max(a)_j}
\end{equation}
\begin{equation}
\text{MultiTask}^{human}_{i} = \frac{1}{m} \sum_{j=1}^{m} \frac{a_{ij} - Human_{j} }{Human_j}
\end{equation}

Greater values of the MultiTask measure indicate better performance across tasks and measures. 

\subsection{Model Comparisons}
Table \ref{tab:unified_human_baselines} shows the performance of the  models on all of the test datasets. 
IMIC-trained foundation models appear in the top section, zero shot models in the second section, and domain experts and humans in the third section.

Given that domain experts are directly tuned for a single task, we expected them to perform better than foundation with IMIC training  and 0-shot foundation models.
This was the case for the cross-pose LFW dataset---a HQ face test.
PetalFace \cite{narayan2024petalface} achieved the maximum performance for CP-LFW, but performed comparably to CLIP-IMIC for LFW.
EVA-02-IMIC and CLIP-IMIC approached domain expert performance on LFW.  
Surprisingly, PetalFace, which was trained to specialize in low resolution face recognition, only exceeded EVA-02-IMIC-A for TinyFaces at Rank 1, with EVA-02-IMIC-A performing better at higher ranks. 
For person recognition, IMIC-trained foundation models surpassed the domain expert for person identification (ECHO-BID \cite{metz2025earlybirdidentifiesworm})
for both the Market-1501 dataset and the more challenging BTS.  


Foundation models in zero-shot tests performed poorly on face and person recognition. As expected, they did well on
object recognition. However, they  
 did not surpass EVA-02-IMIC-A. Moreover, several IMIC-trained models performed at comparable level to the EVA-02 0-shot foundation model.
 This suggests that multi-task training did not interfere with object recognition. 

 For CLIP and DINOv3, fine tuning using either variant of the IMIC curriculum improved performance across all tasks. This held true even for out of distribution tasks such as classification of ImageNet-A. For EVA-02, both IMIC curricula allow for dramatic performance improvements on person recognition tasks, while maintaining best in class performance on object classification. 
 
\begin{figure*}[htbp]
  \centering
  \includegraphics[width=0.45\textwidth]{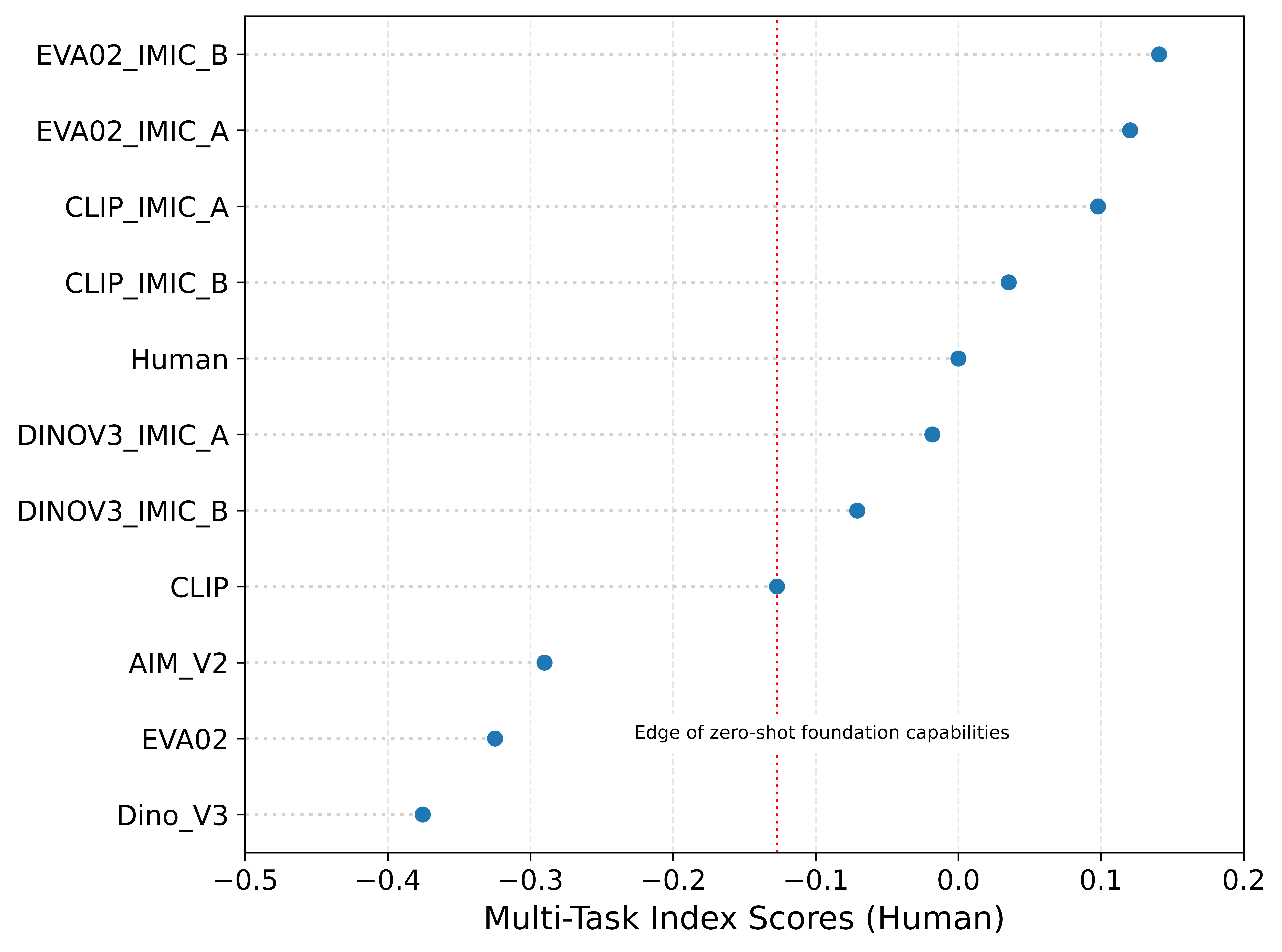}
  \hspace{0.02\textwidth} 
  \includegraphics[width=0.45\textwidth]{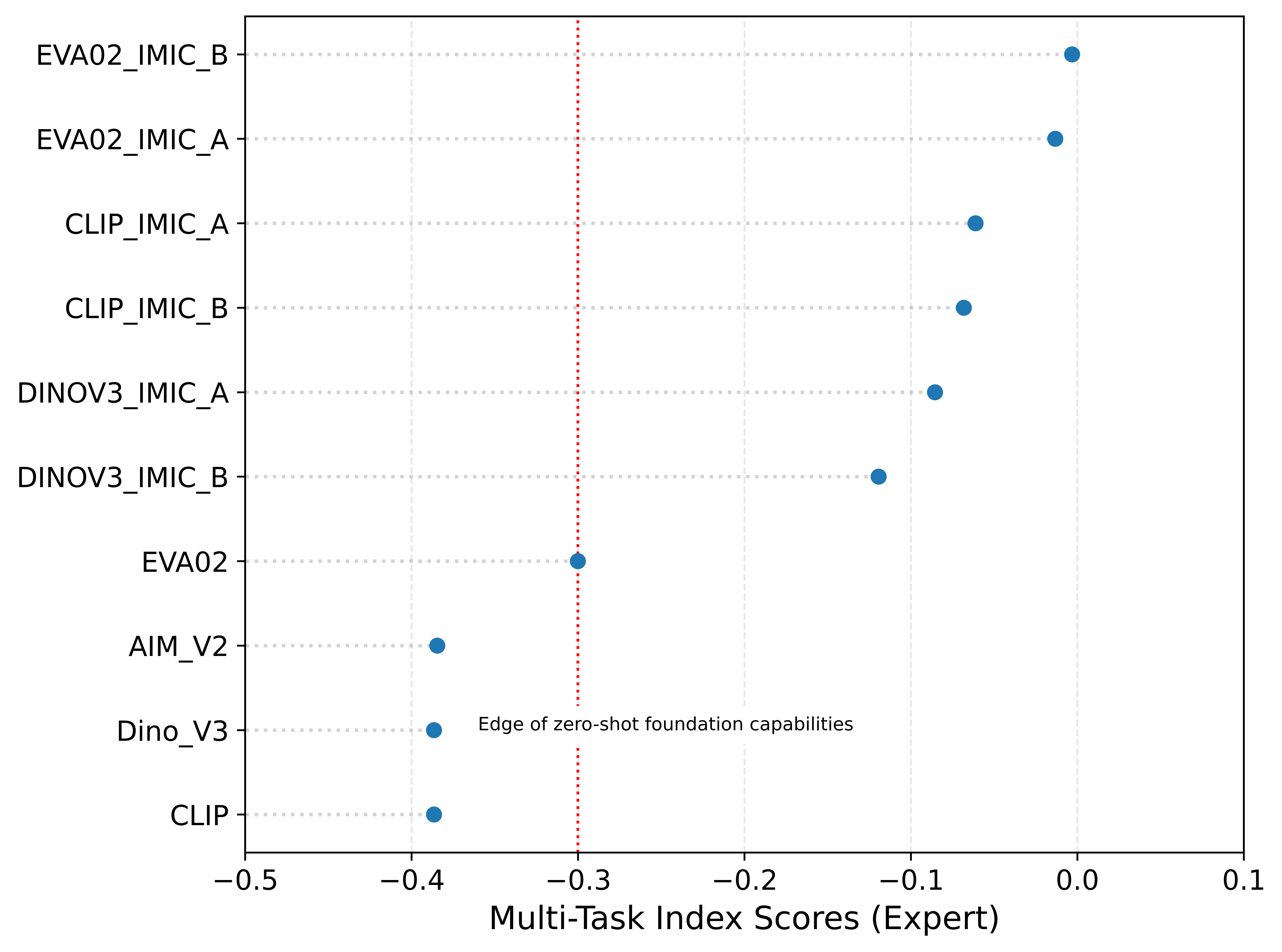}
  \caption{Human Multi-task Index compares of model multi-tasking performance with human multi-tasking performance (left). Expert Multi-task Index shows comparisons with the expert domain models (right). IMIC models are the first models to surpass humans on face, body, and object classification concurrently. These models approach parity with domain experts.}
  \label{fig:multiTaskIndex}
\end{figure*}

\subsection{Multi-tasking}
Figure \ref{fig:multiTaskIndex} shows the multi-task index for all models tested on the full range of tasks. Specifically, this figure provides a clearer picture of cross-task performance. It measures the performance  of zero-shot foundation models and foundation models with IMIC training, {\it relative to the performance of the domain experts}.  EVA-02-IMIC-B is the best multi-tasking model, followed closely by EVA-02-IMIC-A. All other IMIC-trained foundation models multi-tasked substantially better than 0-shot foundation models. Notably, EVA-02-IMIC-B nearly achieves parity with domain experts, while performing all tasks simultaneously. 

\vspace{-0.5 em}
\subsection{Comparison to Human Performance}
Figure \ref{fig:multiTaskIndex} shows the human-centered multi-task index for datasets and measures with relevant
human benchmarks. 
For HQ face identification, we had two human benchmarks: one \cite{6996296} for Labeled Faces in the Wild (LFW) \cite{huang2008labeled} and  one 
for Cross Pose LFW (CP-LFW) \cite{zheng2018cross}. Both measure performance as  $\%$ correct.  For whole-person recognition, we used  a Rank 1 human benchmark \cite{5728825} on Market-1501 \cite{zheng2015scalable}. For object recognition, we used a human benchmark \cite{zhang2018alignedreidsurpassinghumanlevelperformance} on ImageNet-1K \cite{russakovsky2015imagenet} for Rank 5 accuracy.
Algorithms that surpass human multi-task performance score above 0. 
Both EVA-02 and CLIP with IMIC training 
exceed human multi-task performance.
DINOv3 with IMIC training does not perform as well as humans. 
All out-of-the-box foundation models fell well short of human multi-task performance.

In summary, 
these are the first models that surpass human performance on object, face, and whole person identification in a unified embedding space.

\vspace{-0.5 em}

\subsection{Catastrophic Forgetting}
\begin{figure}[htbp]
    \centering
    \includegraphics[width=0.5\textwidth]{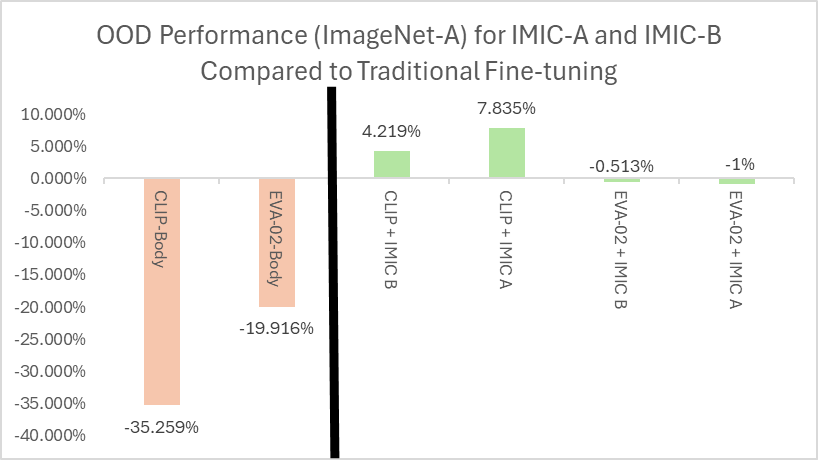}
    \vskip .15 cm
    \caption{Interleaved training can offset catastrophic forgetting. Interleaved models exhibit substantially smaller performance drops on out of distribution (OOD) object classification data as compared to a model trained with body data. Two of the models even   improved at object classification with interleaved multi-task training.}
    \label{fig:OOD_Decline}
\end{figure}

In the main results section, we demonstrate that the IMIC curriculum enables foundation models to be adapted for face and body identification tasks without compromising their performance on object classification. However, we tested object classification using 
the ImageNet validation set, which is an in-distribution dataset with the ImageNet-1k training data. Therefore, to further assess the generalization of IMIC-trained models using out-of-distribution
(OOD) benchmarks, here we test with  ImageNet-A. We trained two control models, CLIP-Body and EVA-02-Body. These control models are traditionally fine-tuned on just the body data utilized in the IMIC curricula. Figure \ref{fig:OOD_Decline}
 shows that models trained with IMIC exhibit performance gains or only minor performance declines on ImageNet-A, in contrast to direct fine-tuning approaches, which lead to substantial performance drops on this dataset. These results indicate that IMIC training enables multi-task improvement while maintaining the desirable generalizability of foundation models.
\vspace{-2 em}
\section{Analysis of Embedding Space}
We have demonstrated the utility of the IMIC training curriculum for simultaneously learning multiple diverse tasks with fundamentally different task requirements, and we show that there is no real forgetting of the original training for objects.  
Next, to determine the characteristics of the 
representation that achieves this feat, we  perform extensive analyses of the embedding space for the EVA-02 model variants.

\subsection{Linear Separability of Tasks}
Do the representations of the four tasks inhabit different parts of the space?
We sampled 10,000 embeddings from each of the four tasks, excluding some classes from ImageNet-1k  that could be confused with human identification tasks (e.g., sweatshirt). We trained a linear classifier to predict the task
(HQ face, LQ face, person, object) from which each embedding originated. Eva-02, Eva-02 + IMIC-A, and Eva-02 + IMIC-B all produced  embeddings that could be separated linearly with high accuracy. Ten-fold cross validation yielded accuracies of $99.9 \pm 0.05$, $99.9 \pm 0.03$, and $99.9 \pm 0.02$, respectively.

\begin{figure}[htbp]
  \centering
   \includegraphics[width=0.45\textwidth]{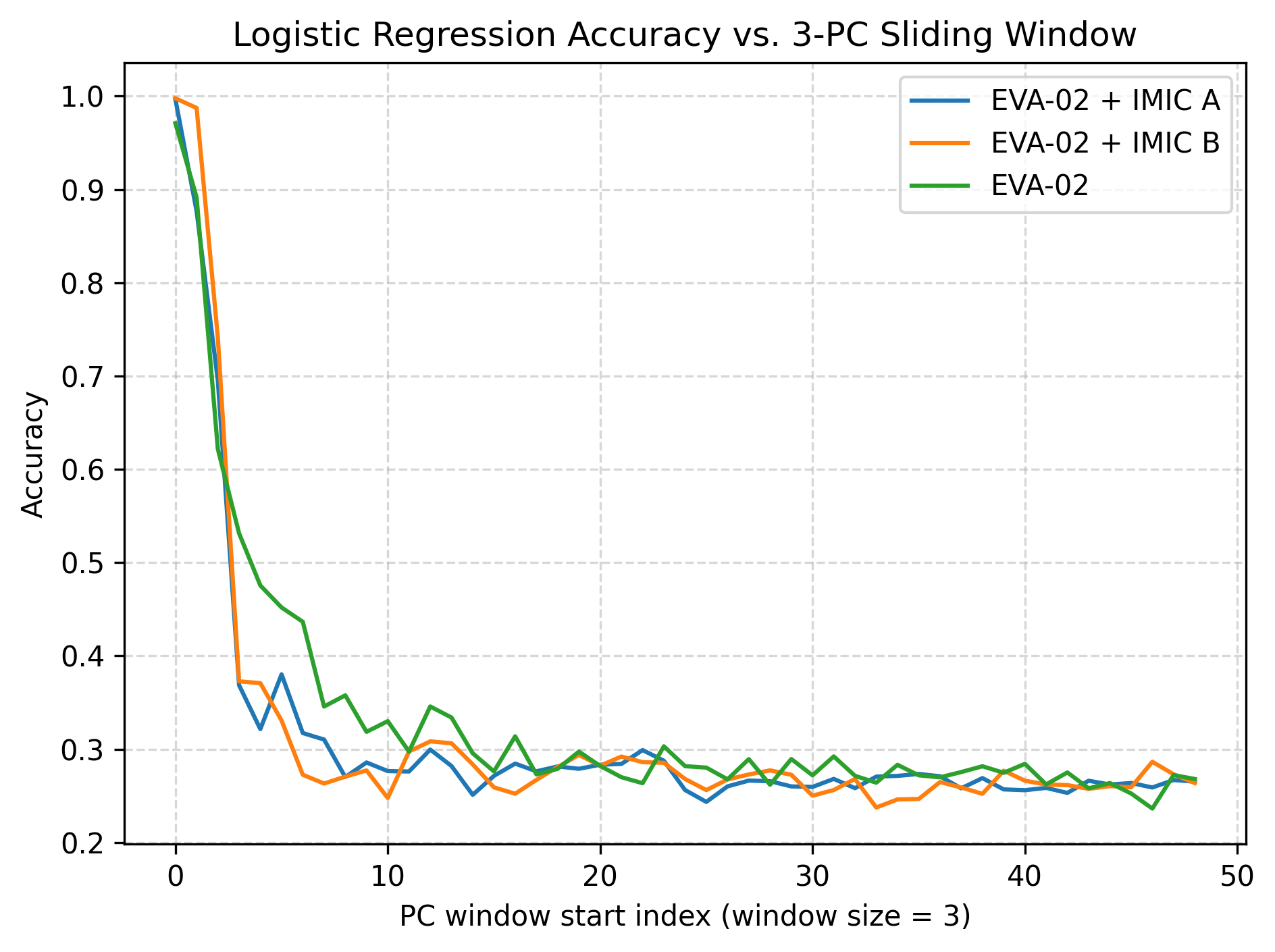}
    \caption{Task separation accuracy as a function of PCs isolating subspaces with a sliding window that moves from early to later PCs.  Early PCs are highly effective at task separation.}
    \label{fig:PC Seperation}
\end{figure}

We repeated this analysis in the principal component (PC) space to assess task separation in lower-dimensional representations. Figure \ref{fig:PC Seperation} shows the accuracy of task separability with a sliding window of 3 PCs. 
Task separability is highly accurate with only the first three PCs. As the sliding window moves to later PCs, task separation accuracy decreases markedly. The figure indicates clearly that task-discriminative information is concentrated in the early PCs.

\begin{figure}[htb]
\begin{tabular}{ll}
\includegraphics[width = .48 \columnwidth]{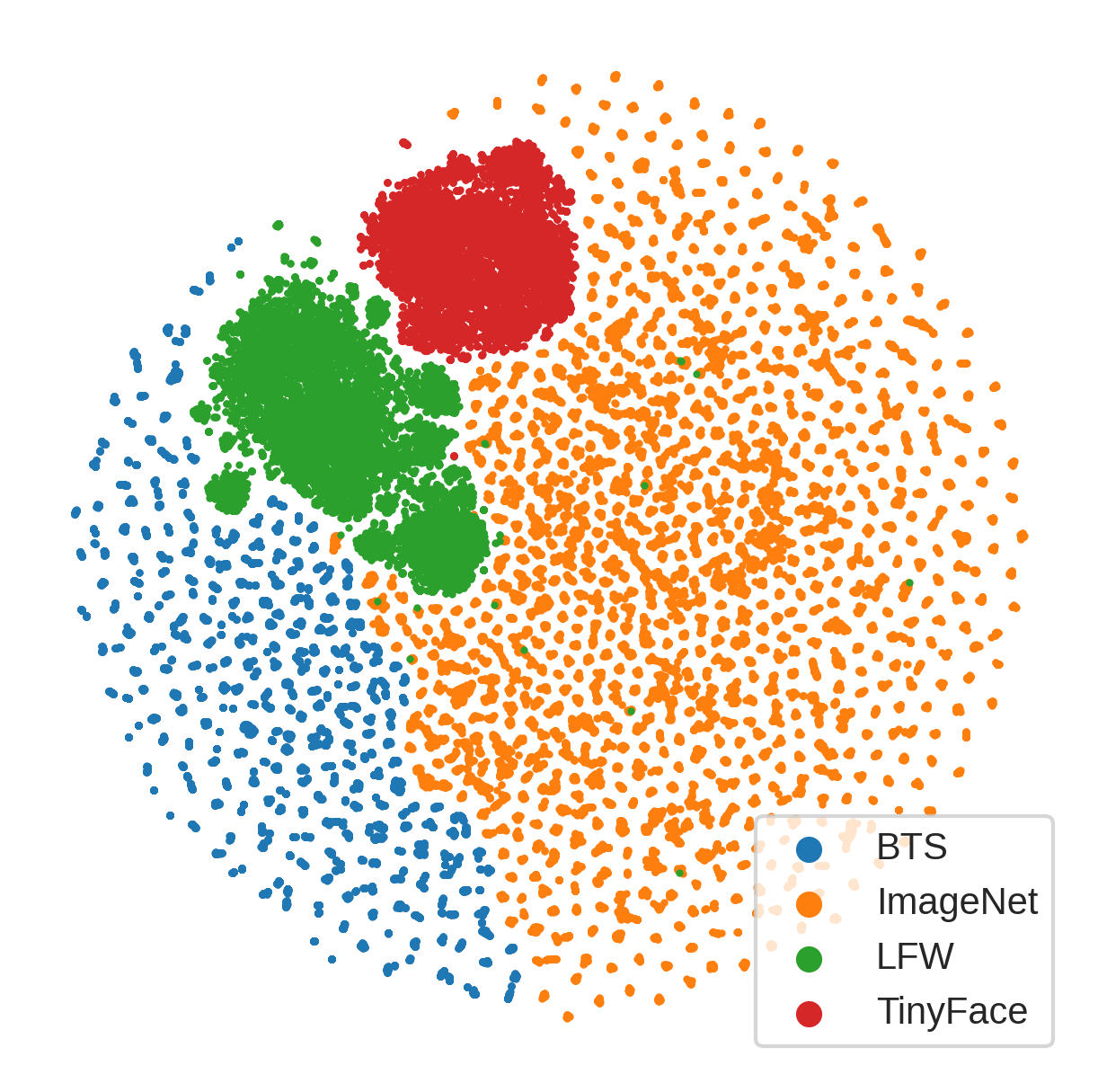}
&
\includegraphics[width = .48 \columnwidth]{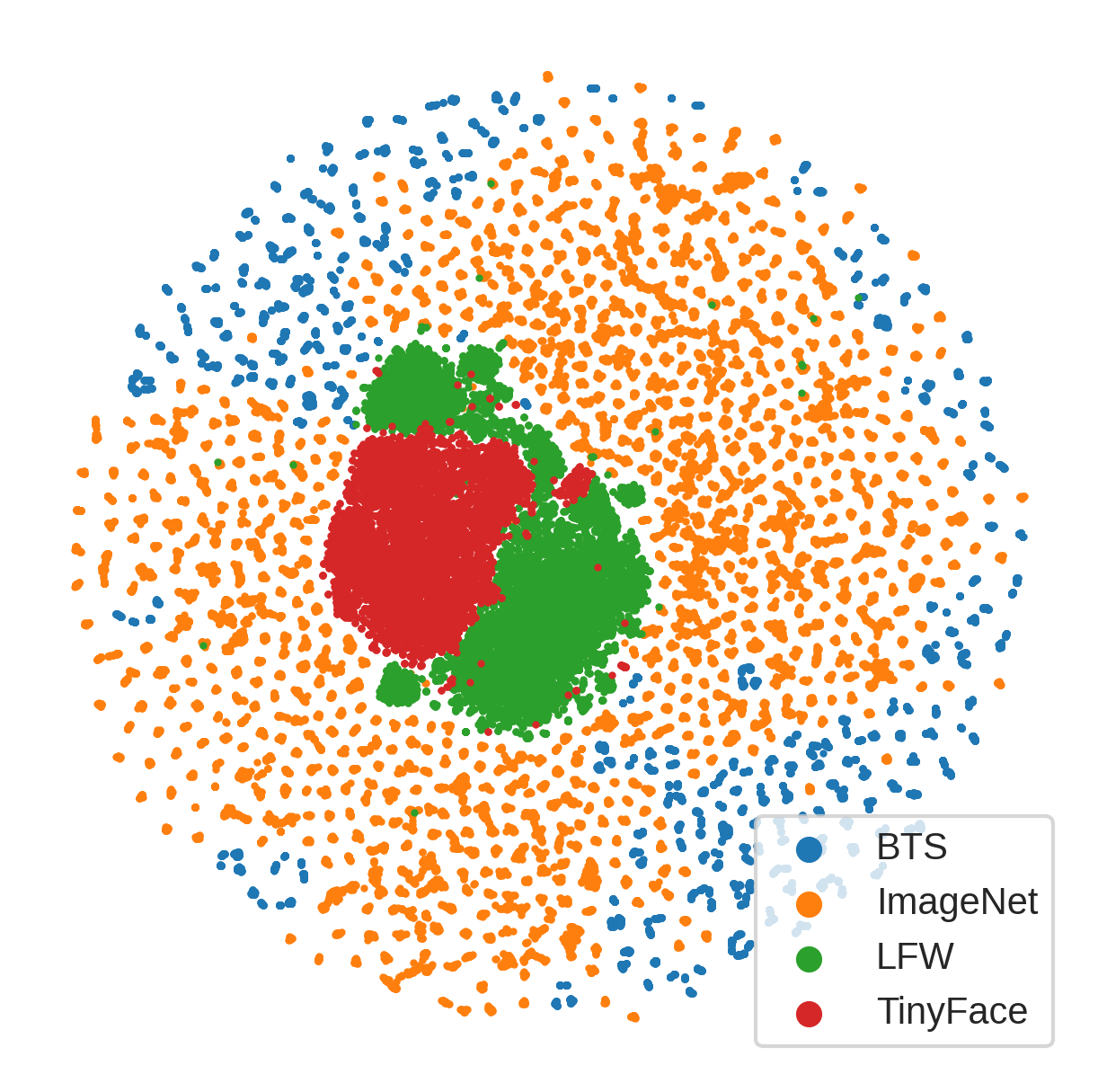}
\end{tabular}
\caption{$t$-SNE visualization of raw embeddings generated from BTS (body identification), ImageNet (object classification), LFW (HQ face identification), and TinyFace (LQ face identification). Left: PCs 1--800. Right: PCs 4--800. Perplexity = 30 in both plots. Perplexity values ranging from 10 to 1,000 showed no difference in overall task separability. }
  \label{fig:tsne}
\end{figure}

\vspace{-0.5 em}
\subsection{t-SNE Visualization}

Next, we created dimensionality reduction visualizations of raw embeddings using $t$-distributed Stochastic Neighbor Embedding ($t$-SNE) \cite{maaten2008visualizing}. These  appear 
in Figure \ref{fig:tsne}. The left figure shows the subspace  defined by PCs 1--800.  Task representations are 
clearly separated. 
The right figure shows the subspace defined by PCs 4--800 (i.e., excluding the PCs that account for the majority of task separation). The \mbox{$t$-SNE} is consistent with feature sharing among the representations, such that the embeddings show a nested structure whereby faces of high and low quality are embedded within  a larger space defined  by objects and bodies.

\begin{figure}[t]
   \centering
     \includegraphics[scale = .40]{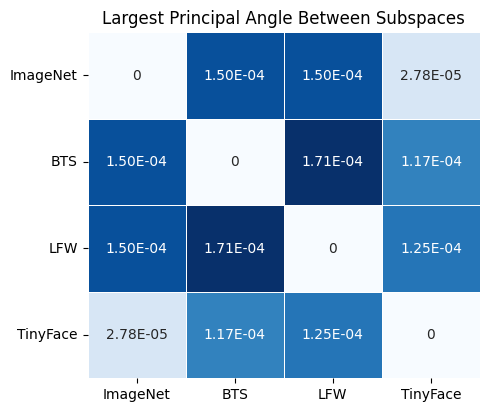}
     \vskip 0.15 cm
   \caption{For EVA-02 models + IMIC, principal angles between subspaces constructed using embeddings from each task are very low, regardless of chosen loss function or compared dataset. This provides evidence that interleaved fine-tuning promotes features which are strongly shared across the four tasks. Shown are the largest angles for EVA-02 + IMIC-A}
   \label{fig:principal-angles}
 \end{figure}
\vspace{-0.5 em}
\subsection{Feature Subspaces}
Linear separability of task-specific embeddings, even combined with visualization, provides only preliminary information about structure of the space. Linear separability ensures that embeddings from different tasks occupy distinct regions of the embedding space, but it offers limited insight into the extent to which features are shared across tasks. Because the geometric structure of the embedding space is invariant under orthogonal rotations, we propose  an additional approach for assessing feature sharing by comparing the directions along which task-specific embeddings vary.

Given two subspaces of $\mathbb{R}^m$, $F$ and $G$, principal angles $\theta_k$ can be recursively defined as: 
\[
\cos \theta_k \;=\; 
\max_{\substack{u \in F, \, v \in G}} 
u^\top v, \|u\|_2 = 1, \, \|v\|_2 = 1
\] under the constraint that $u^\top u_j = 0$ and $v^\top v_j = 0$ for $j = 1,2, ..., k-1$.

Intuitively, principal angles attempt to find unit vectors in $F$ and $G$ that are maximally aligned. If $dim(F) \leq dim(G)$, principal vectors form an orthonormal basis for $F$ and span a subspace of $G$. A principal angle $\theta_k = 0$ indicates that the two subspaces intersect in one direction, with the count of 0 angles dictating the number of shared directions. A principal angle $\theta_k = \frac{\pi}{2}$ indicates that at least one direction in the subspaces is orthogonal, with the count of $\frac{\pi}{2}$ angles dictating the number of orthogonal directions. 

In this analysis, let $\mathbb{R}^{1568}$ be the full possible embedding space. Define task-specific subspaces as the span of the number of principal components required to explain $99\%$ of task-specific variance. The principal angles between these subspaces are then calculated.

Fig \ref{fig:principal-angles} shows the principal angles between subspaces for the high quality faces, low quality faces, whole people, and objects.  The largest principal angles are all smaller than .001, indicating features that align between subspaces.
In summary, despite being
linearly separable, the subspaces defined by each task
share features.   
\vspace{-0.5 em}
\subsection{Feature Reconstruction in Subspace}


Principal vectors between two subspaces offer a basis only up to $min(dim(F), dim(G))$. As such, this analysis cannot tell us whether features that occupy the subspace $F$ or $G$ are sufficient to perform recognition for each class of items. Although it is well known that few PCs from a full embedding space can reproduce performance accurately, it is unintuitive to expect PCs from a specific task to work for another task, even if these tasks are represented in a single space.
Therefore, we next asked  whether the feature dimensions (PCs) of a given task (i.e., the subspace in which task items are represented) can be used to perform the other tasks.

We  defined task-specific subspaces by taking the embedding matrix from a test set used in each task (e.g., LFW for HQ faces) and performing a PCA on these embeddings. We then tested performance with a sliding subset of PCs as the basis of the associated task (i.e., subspace of unified embedding space). Next, for each other task, we projected test embeddings from that task into this subspace, and used these projections to test classification performance. 

\begin{figure}[t]
   \centering
   \includegraphics[width = .9 \columnwidth]{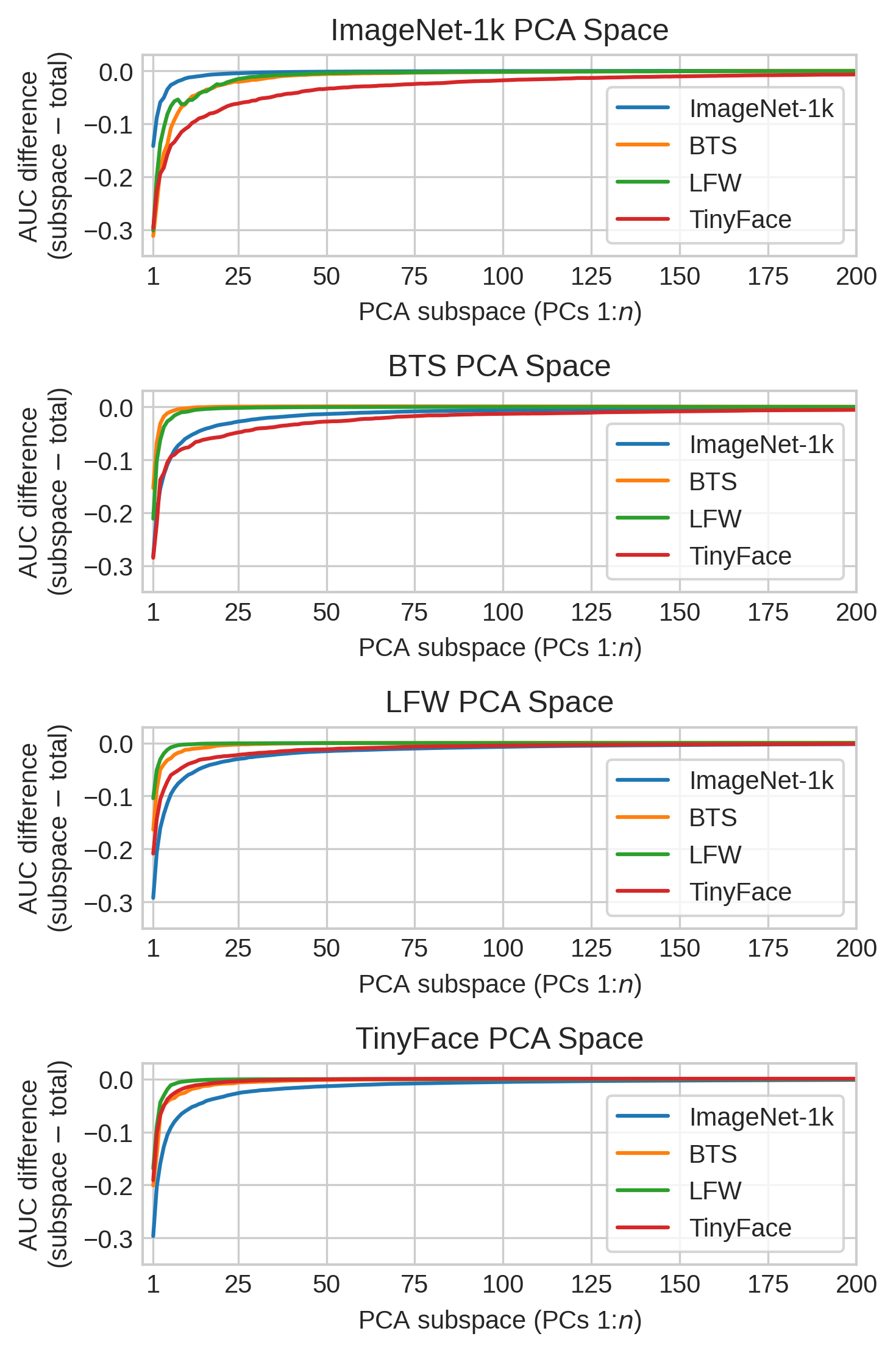}
  \caption{For each PCA basis (top to bottom panel: ImageNet-1k, BTS, LFW, TinyFace), performance differences between full-dimensional embeddings and subspaces from 1D to 200D. Using only 100 principal components fit to embeddings from a different task, a maximum performance loss $<.02$ AUC (TinyFace performance in 100D Imagenet PCA subspace). Rank 1 and Rank 20 accuracy show similar patterns.}
  \label{fig:subspaces}
\end{figure}

Results appear in Figure \ref{fig:subspaces} and can be summarized as follows. The basis features (PCs) of each task subspace are capable of performing all of the other recognition tasks with only minimal loss in accuracy ($<.02$ AUC) by comparison to accuracy in the full embedding space. Slightly larger numbers of PCs were required when the basis set differs from the task being tested.

In summary,
this analysis shows that the principal component embedding basis from any individual task subspace can perform all of the other tasks. Only $\approx$100 PC dimensions from any task-specific subspace were necessary to yield nearly full performance on all other tasks.


\vspace{-1 em}
\section{Conclusions}
\vskip -0.15 cm
We began with an open question in visual neuroscience: How does the brain learn highly specialized representations of faces, bodies, and objects from a diverse and variable input stream? This problem is further complicated by the differing requirements of intra- and inter-class discrimination.  
The models we introduce achieved face, body, and object identification at levels that surpassed human multi-tasking and multi-tasking by 0-shot foundation models.
Our interleaved training curriculum accomplished this feat, by partitioning
the ``unified'' embedding space into task-specialized regions along the first 3 PCs. This outcome mirrors the domain-specific face, body, and object areas in the human visual cortex \cite[cf.][]{grill2014functional}. 
Beyond that, converging data from our analyses indicate that the linearly separable domain-specific regions share 
feature dimensions across tasks in the higher dimensions of the space. Feature sharing across domains also occurs in human visual cortex \cite{vinken2023neural}. 
In sum, this work provides empirical evidence that a single model can encompass representations that support  face, person, and object identification in  both domain-general and domain-specific ways.

\section{ACKNOWLEDGMENTS}

This research is based upon work supported in part by the Office of the Director of National Intelligence (ODNI), Intelligence Advanced Research Projects Activity (IARPA), via [2022-21102100005]. The views and conclusions contained herein are those of the authors and should not be interpreted as necessarily representing the official policies, either expressed or implied, of ODNI, IARPA, or the U.S. Government. The U.S. Government is authorized to reproduce and distribute reprints for governmental purposes notwithstanding any copyright annotation therein.

{
\small
\bibliographystyle{ieeenat_fullname}
\bibliography{main}
}


\end{document}